\documentclass[10pt,twocolumn,letterpaper]{article}

\usepackage{dicta}
\usepackage{times}
\usepackage{epsfig}
\usepackage{graphicx}
\usepackage{amsmath}
\usepackage{amssymb}
\usepackage{tabularx, booktabs}
\usepackage{algorithm}
\usepackage{algpseudocode}
\usepackage{pifont}
\usepackage{cite}
\newcommand{\cmark}{\ding{51}}
\newcommand{\xmark}{\ding{55}}
\usepackage[table]{xcolor}
\usepackage{multirow}
\newcommand{\coolname}{\textit{REDAR}}

\usepackage[pagebackref=true,breaklinks=true,letterpaper=true,colorlinks,bookmarks=false]{hyperref}

\dictafinalcopy 

\def\dictaPaperID{64} 

\ifdictafinal\fi
\begin{document}


\title{REDAR: Reinforcement-Driven Adaptive Regularisation for Class Incremental Learning
}


\author{
Akila Ayanthi$^{1,2}$, Darshana Priyasad$^{1}$, Tharindu Fernando$^{1}$, Sridha Sridharan$^{1}$, \\ Clinton Fookes$^{1}$, Peyman Moghadam$^{1,2}$ \\
{\small $^{1}$SAIVT, Queensland University of Technology, Australia \quad
$^{2}$CSIRO Robotics, CSIRO, Australia} \\
{\small \texttt{\{a.mudiyanselage, dp.don, t.warnakulasuriya, s.sridharan, c.fookes\}@qut.edu.au}} \\
{\small \texttt{peyman.moghadam@csiro.au}}
}

\maketitle
\begin{abstract}
Regularisation-based class-incremental learning (CIL) methods control the stability-plasticity trade-off through fixed coefficients tuned offline, independently of the continual learning objective and blind to phase-specific variations in data volume, task difficulty, and class similarity. We argue this is a fundamental mismatch and propose \textbf{\coolname{}}: \textbf{Re}inforcement-\textbf{D}riven \textbf{A}daptive \textbf{R}egularisation for Class Incremental Learning, which treats regularisation strength selection as a sequential decision problem. At each incremental phase, a lightweight reinforcement learning (RL) agent observes a compact state of current performance metrics and incoming data statistics, and outputs the distillation coefficients for that phase. To respect CIL's constraint on historical data access, the policy is trained entirely on pseudo-CIL tasks constructed from the initial phase, then deployed on the incremental sequence without further adaptation. Experiments on CIFAR-100 and Tiny-ImageNet demonstrate that \coolname{} reduces forgetting and improves forward transfer over fixed and heuristic baselines, and that policies trained on short pseudo-task horizons transfer effectively to longer incremental runs. 
\end{abstract}

\section{Introduction}
\label{sec:intro}

Class-incremental learning (CIL) requires a model to absorb new categories sequentially while retaining performance on all previously seen classes, without access to the full historical dataset. This capability is essential for deployed perception systems and embodied agents that operate under non-stationary data streams. The central challenge is catastrophic forgetting: naive fine-tuning overwrites earlier representations as the network adapts to new data, exposing the fundamental stability–plasticity dilemma.

Prior work addresses this from three angles: rehearsal, which interleaves stored exemplars with current training data \cite{RebuffiICaRL:Learning, LiuRMM:Learning}; architecture-based methods, which grow or partition capacity to reduce interference \cite{Yan2021DER:Learning, Luo2023Class-IncrementalLearning}; regularisation-based methods, which constrain updates via parameter penalties \cite{KirkpatrickOvercomingNetworks, Zenke2017ContinualIntelligence} or knowledge distillation \cite{Li2018LearningForgetting, HouLearningRebalancing, Douillard2020PODNet:Learning}. 
Despite this progress, one design choice has received little attention: how strongly should a model be regularised at each incremental step? Existing methods fix the regularisation coefficient across all phases, selected once via offline hyperparameter search \cite{RebuffiICaRL:Learning, HouLearningRebalancing, Douillard2020PODNet:Learning, FENG2024103629, roy2023subspace}. 
This is a poor fit for realistic CIL scenarios, where successive phases differ in data volume, class imbalance, task difficulty, and similarity to prior classes, so the stability--plasticity trade-off shifts in ways a single global coefficient cannot track.

We treat regularisation strength as a control variable and formulate its selection as a sequential decision problem. To this end, we propose \textbf{\coolname{}} (\textbf{Re}inforcement-\textbf{D}riven \textbf{A}daptive \textbf{R}egularisation for Class Incremental Learning), which casts per-phase coefficient selection as a policy learned via Reinforcement Learning (RL). 
A lightweight RL agent observes a compact state consisting of current performance and basic data statistics, and outputs a per-phase distillation coefficient. To respect CIL's constraint on historical data access, we train the policy offline on pseudo-CIL tasks constructed from the initial phase data alone, then deploy it on the true incremental sequence without retraining, using distributed experience collection.

We make three empirical claims. First, a state-dependent learned policy reduces forgetting and improves forward transfer relative to fixed and heuristic baselines on CIFAR-100 and Tiny-ImageNet (Sec.~\ref{sec:acc_vs_frg}). Second, policies trained on short pseudo-task horizons transfer to longer incremental runs (Sec.~\ref{sec:cross_task}). Third, the gains are attributable to learned state-conditioned selection: the policy outperforms both the best fixed coefficient and random per-phase draws from the same action space (Sec.~\ref{sec:adapt_reg}).

\section{Related Work}
\label{sec:related_work}

\subsection{Class-Incremental Learning}
\vspace{-2mm}
Existing CIL methods fall into three broad categories. \textbf{Replay-based methods} maintain a small exemplar buffer or synthesise past data, interleaving stored samples with current training~\cite{RebuffiICaRL:Learning,ChaudhryOnLearning,Lopez-PazGradientLearning}. When real data cannot be reused, synthetic images or features are replayed instead~\cite{ShinContinualReplay,XiangIncrementalNetworks}, while exemplar-free variants replay class prototypes and compensate for their drift~\cite{peng2024multivariate}. \textbf{Architecture-based methods} grow or route capacity to reduce inter-task interference~\cite{Luo2023Class-IncrementalLearning,FENG2024103629}, but become impractical as the number of tasks grows. \textbf{Regularisation-based methods} add penalty terms to prevent drift. Weight-based approaches penalise changes to important parameters~\cite{KirkpatrickOvercomingNetworks,Douillard2020PODNet:Learning,Zhao2020MaintainingLearning}, while activation-based approaches distill knowledge from the previous model at the logit~\cite{Li2018LearningForgetting}, feature~\cite{HouLearningRebalancing}, or embedding-geometry level~\cite{roy2023l3dmc}. Flashback learning~\cite{MAHMOODI2025107616} makes regularisation bidirectional, pulling toward both old and new knowledge. Most of these methods use fixed regularisation strengths~\cite{RebuffiICaRL:Learning,Buzzega2020DarkBaseline} or hand-crafted schedules~\cite{HouLearningRebalancing,Douillard2020PODNet:Learning}. Recent work adapts regularisation via Bayesian optimisation~\cite{Gok2023AdaptiveRegCIL} or prompt-based regularisers on pre-trained backbones~\cite{Huang2024OVOR}, but treats each phase in isolation without a long-horizon objective. \coolname{} addresses this gap by casting coefficient selection as a sequential decision problem, accounting for the downstream effects of each phase's regularisation strength.


\subsection{Reinforcement Learning for CIL}
\vspace{-4pt}

RL has achieved strong results in robot control~\cite{Ibarz2021HowLearned}, game playing~\cite{Silver2016MasteringSearch}, and neural architecture search~\cite{Zoph2016NeuralLearning}, and has been applied to vision problems such as active learning~\cite{CaicedoActiveLearning} and object detection~\cite{Kotar2022Interactron:Detection}. In continual learning, it has been used to optimise specific design choices: RCL~\cite{XuReinforcedLearning} learns to expand network capacity for new tasks, and RMM~\cite{LiuRMM:Learning} learns a memory allocation policy across phases, extended to select which samples~\cite{Zhu2023ContinualSelection} or tasks~\cite{KlassonLearnLearning} to replay. Most closely related, \cite{Liu2023OnlineLearning} adapts regularisation strength as an online bandit problem, but must be re-optimised per phase and does not condition on a rich phase description; \cite{Gok2023AdaptiveRegCIL} similarly tunes regularisation independently at each step via Bayesian optimisation.

In contrast, \coolname{} formulates regularisation selection as a sequential decision problem with an explicit state of performance metrics and data properties. Trained offline on pseudo-CIL tasks and deployed without retraining, it optimises a long-horizon CIL objective and generalises across incremental horizons.
\section{Preliminaries}
\label{sec:preliminaries}

\subsection{Class Incremental Learning}
\vspace{-4pt}



Class incremental learning (CIL) spans $(N+1)$ phases: an initial phase followed by $N$ incremental phases, with the class count growing at each stage.


\noindent\textbf{Initial Phase.} The model $\theta_0$ is randomly initialised and trained on $D_0$, comprising all samples from the first 50\% of classes (\textit{train-from-half}). A fixed-size memory buffer $\mathcal{M}$ (budget $K_{mem}$) is populated via herding, selecting exemplars per class from $D_0$.

\noindent\textbf{Incremental Phases.} In phase $i$, $\theta_i$ is initialised from $\theta_{i-1}$ and trained on new classes $D_i$ combined with memory $\mathcal{M}$, then fine-tuned on a class-balanced subset of all observed classes to correct recency bias from the new-memory data imbalance. After training, $\mathcal{M}$ is updated via herding to retain $K_{mem} / |C_{0:i}|$ exemplars per class, where $C_{0:i}$ is the set of all classes seen up to phase $i$.






\section{Methodology}
\label{sec:method}
\vspace{-4pt}

\begin{figure*}[ht]
    \centering
    \includegraphics[width=0.95\textwidth]{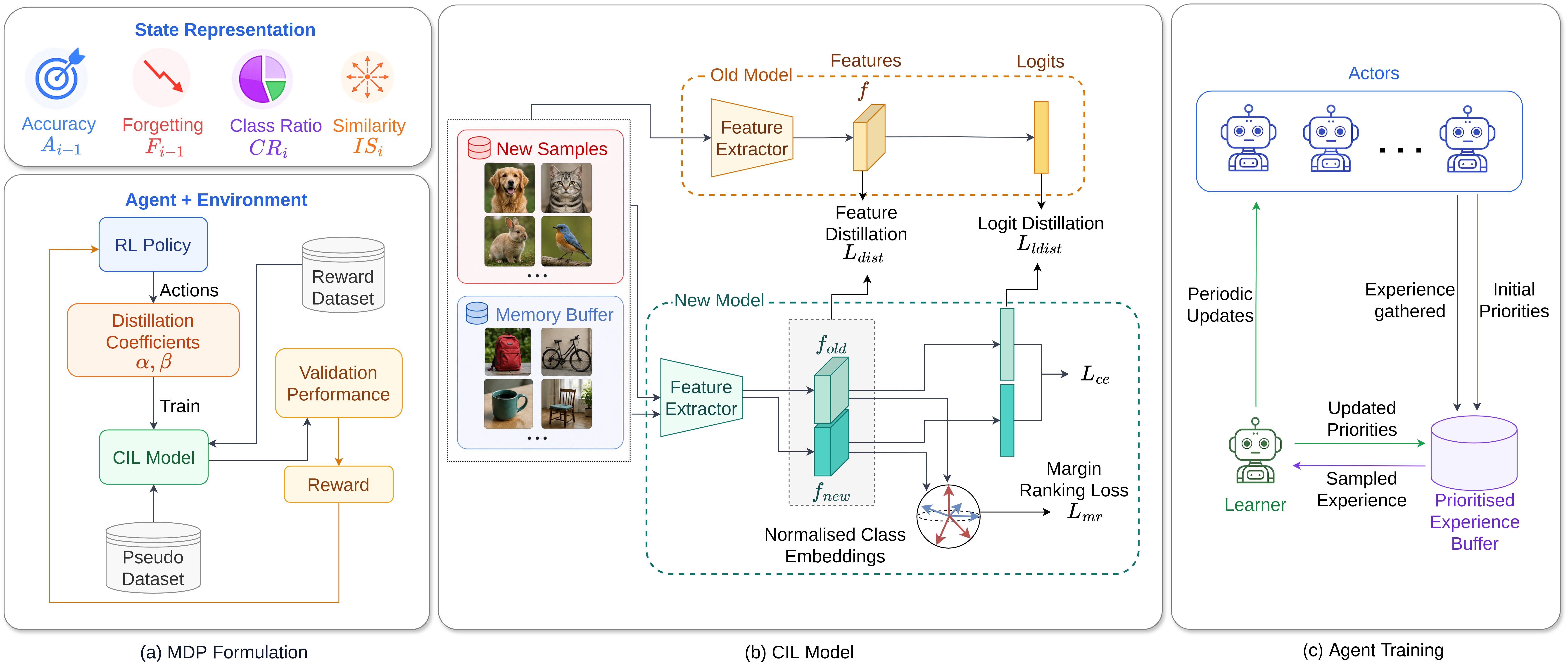}
     \caption{\textbf{Overview of \coolname.}
    \textbf{(a)} Reinforcement-driven regularisation is formulated as a finite-horizon Markov Decision Process (MDP), $\mathcal{M}=(\mathcal{S},\mathcal{A},\mathcal{T},\mathcal{R},\gamma)$, where the state $s_i=(A_{i-1},F_{i-1},CR_i,IS_i)$ summarises the current learning dynamics, the action $a_i=(\alpha_i,\beta_i)$ specifies the phase-specific regularisation coefficients, and the reward $r_i$ is the validation accuracy after CIL training.
    \textbf{(b)} The CIL model jointly optimises cross-entropy, feature distillation, logit distillation, and margin ranking losses while learning from replay memory and incoming samples.
    \textbf{(c)} The policy is trained offline on pseudo-CIL tasks using a distributed actor--learner architecture with prioritised experience replay before being deployed on the target incremental sequence.}
    \label{fig:flow}
\vspace{-2mm}
\end{figure*}

We propose \coolname{} (Figure~\ref{fig:flow}), an RL framework for adaptive regularisation in CIL with two components: a CIL model parameterised by phase-specific distillation coefficients ($\alpha_i, \beta_i$), and an RL agent that learns to set them. 
We describe the CIL model in Section \ref{sec:cil_model}, formulate coefficient selection as an MDP in Section \ref{sec:rl_reg}, and detail agent training on pseudo-tasks in Section \ref{sec:agent_training}.

\subsection{CIL Model}
\label{sec:cil_model}
\vspace{-4pt}


The CIL model $f_{\theta,w}: \mathcal{X} \rightarrow \mathbb{R}^K$ is a deep convolutional network composed of a feature extractor $h_{\theta}: \mathcal{X} \rightarrow \mathbb{R}^d$ and a cosine-normalised classifier $g_w: \mathbb{R}^d \rightarrow \mathbb{R}^K$ \cite{HouLearningRebalancing}, given by $f_{\theta,w}(x) = \cos(g_w, h_\theta(x))$. Training combines four losses. First, cross-entropy over all observed classes provides the classification objective: 
\vspace{-2mm}
\begin{align}
L_{ce}(x) = -\sum_{k=1}^{K} y_k \log P_{w_k}(x),
\end{align}
\vspace{-2mm}

where $P_w(x)$ denotes the softmax probabilities from the cosine-normalised classifier. Then, to preserve prior knowledge, we apply distillation in both feature and logit space. The feature distillation loss penalises drift in representation orientation:
\vspace{-2mm}
\begin{align}
L_{fdist}(x) = 1 - \cos\!\left(h_{\theta}^i(x),\, h_{\theta}^{i-1}(x)\right),
\end{align}
and the logit distillation loss penalises divergence in output distributions over previously seen classes (with cosine normalisation omitted to preserve logit scale):
\begin{align}
\mathcal{L}_{ldist}(x) = D_{KL}\big(\hat{p}^{\,i-1}(x)\,\|\,\hat{p}^{\,i}(x)\big),
\end{align}
where $\hat{p}_w^{\,i}$ and $\hat{p}_w^{\,i-1}$ denote the softmax distributions of the current and previous models restricted to the $K_{i-1}$ previously seen classes.

Finally, a margin ranking loss separates old-class exemplar embeddings from new-class embeddings by a margin $m$:
\begin{align}
\mathcal{L}_{mr}(x) = \sum_{h=1}^{H} \max\big(m - z^{y}(x) + z^{(h)}(x),\, 0\big),
\end{align}
where $z^{y}(x)$ is the ground-truth class score, $z^{(h)}(x)$ is the $h$-th hardest new-class score, and $H$ is the number of hard negatives.. The overall training objective at phase $i$ is:

\begin{equation}
\label{eq:overall-loss}
\begin{aligned}
L_{\text{overall}}
=
\frac{1}{|\mathcal{B}|}
\sum_{x \in \mathcal{B}}
\Bigl(
L_{ce}(x)
+
\alpha_i\,L_{fdist}(x)
+
\beta_i\,L_{ldist}(x)
\Bigr)
\\ +
\frac{1}{|\mathcal{B}_{old}|}
\sum_{x \in \mathcal{B}_{old}}
L_{mr}(x),
\end{aligned}
\end{equation}
where $\mathcal{B}$ denotes a training mini-batch drawn from the current phase dataset, and $\mathcal{B}_{old}\subseteq\mathcal{B}$ denotes the subset of replay samples belonging to previously learned classes, and $\alpha_i$ and $\beta_i$ are phase-specific coefficients weighting representation and prediction stability, respectively. Since $L_{ce}$ and $L_{mr}$ are always active, the model continually acquires new classes; adjusting $(\alpha_i, \beta_i)$ controls how strongly it must simultaneously preserve past knowledge. These two coefficients therefore govern the stability-plasticity trade-off at each phase and constitute the action space of our RL agent, described next in Section \ref{sec:rl_reg}.


\subsection{Reinforcement-Driven Regularisation}
\label{sec:rl_reg}
We formulate regularisation selection as a finite-horizon Markov Decision Process (MDP), $\mathcal{M} = (\mathcal{S}, \mathcal{A}, \mathcal{T}, \mathcal{R}, \gamma)$. The environment is the CIL training process: each episode is a full run of $N$ incremental phases, each phase is one environment step, and we set $\gamma=1$. At step $i \in \{0,\dots,N-1\}$, the agent observes a state $s_i \in \mathcal{S}$ summarising current model performance and incoming-phase data statistics, selects an action $a_i \in \mathcal{A}$ corresponding to the distillation coefficients $(\alpha_i, \beta_i)$ for that phase, and receives a scalar reward $r_i = \mathcal{R}(s_i, a_i)$ after the CIL model has been trained under $a_i$. The policy $\pi_\theta(a \mid s)$ is trained to maximise the undiscounted cumulative return $\sum_{i=0}^{N} r_i$. We instantiate $\mathcal{S}$, $\mathcal{A}$, and $\mathcal{R}$ in the following subsections.

\subsubsection{State Representation}
\label{sec:state}
\vspace{-4pt}

The state $s_i$ is constructed by concatenating two groups of features that jointly characterise the factors governing the appropriate regularisation strength at each phase: the model's current learning behaviour and the statistical properties of the incoming data. 

\noindent\textbf{Performance metrics.} We include two scalar indicators. \emph{Average accuracy} after phase $i$ measures overall competence across all tasks seen so far:
\begin{align}
\label{eq:accuracy}
A_i = \frac{1}{i}\sum_{j=1}^{i} A_{i,j},
\end{align}
where $A_{i,j}$ is the accuracy on task $j$ after phase $i$. Since $A_i$ conflates retention of old knowledge with acquisition of new, we also track \emph{average forgetting}. 
The forgetting on task $j$ after phase $i>j$ is defined as the drop from its peak accuracy, 
$F^j_i = \max_{l \in \{j,\dots,i-1\}} A_{l,j} - A_{i,j},\ \forall j < i$. The average forgetting across all prior tasks is:
\begin{align}
\label{eq:avg_frg}
F_i = \frac{1}{i-1}\sum_{j=1}^{i-1} \mathcal{F}_i^{j}.
\end{align}

\noindent\textbf{Data properties.} Performance metrics alone do not capture the demands of the incoming task. We include two data-level cues: class ratio and inter-task similarity.

The \emph{class ratio} quantifies the relative volume of new data at phase $i$:
\begin{align}
\label{eq:cls-ratio}
CR_i = \frac{\sum_{j=0}^{i-1} |C_j|}{|C_i|},
\end{align}
where $|C_j|$ is the number of classes introduced at phase $j$. A large $CR_i$ means the incoming phase adds far fewer classes than those already consolidated, placing greater pressure on stability.

The \emph{inter-task similarity} measures how closely the new classes relate to previously learned ones, informing whether prior representations can be reused or stronger plasticity is needed. We estimate it from class-mean embeddings under a frozen ImageNet-pretrained ResNet-34:
\begin{align}
\label{eq:mean_feat}
\mu_c = \frac{1}{|P_c|}\sum_{p \in P_c} \phi(x_p),
\end{align}

where $P_c$ is the sample set of class $c$ and $\phi(x_p) \in \mathbb{R}^d$ is the embedding of sample $x_p$. The mean distance between incoming and consolidated classes at phase $i$ is:
\begin{align}
\label{eq:avg_dist}
D_i = \frac{1}{|C_i|}\sum_{c \in C_i} \frac{1}{\sum_{j<i}|C_j|} \sum_{j<i}\sum_{c' \in C_j} \lVert \mu_c - \mu_{c'} \rVert_2,
\end{align}
which we convert to a similarity score via $IS_i = \frac{1}{1 + D_i}$ so that higher $IS_i$ indicates greater similarity to prior classes. The pretrained extractor is used solely for this estimate and plays no role in CIL model training. Combining all four cues, the state at phase $i$ is $s_i = (A_{i-1}, F_{i-1}, CR_i, IS_i)$.

\subsubsection{Action Space}
\label{sec:action}
\vspace{-4pt}

The action at phase $i$ is the pair $a_i = (\alpha_i, \beta_i)$ controlling the two distillation coefficients in Eq. \ref{eq:overall-loss}. We discretise the action space into a finite grid, $\mathcal{A} = \Lambda_\alpha \times \Lambda_\beta$, where $\Lambda_\alpha$ and $\Lambda_\beta$ are predefined value sets for each coefficient.
Discretisation yields a compact action space that reduces policy-gradient variance and avoids constrained continuous-action methods, at the cost of slightly coarser regularisation control.

\subsubsection{Reward}
\label{sec:reward}
\vspace{-4pt}

We use per-phase validation accuracy as the reward: $r_i$ is the accuracy of the CIL model on the held-out reward split at the end of phase $i$. The episodic return over $N$ incremental phases is $R = \sum_{i=1}^{N} r_i$.
Maximising $R$ incentivises the agent to sustain strong performance throughout the full learning trajectory rather than trading early-phase retention for higher terminal accuracy.

\subsection{Agent Training}
\label{sec:agent_training}

\subsubsection{Pseudo-Task Construction}
\label{sec:pseudo}
\vspace{-4pt}

Applying RL to CIL requires reward signals at each phase, which are unavailable during standard incremental training since past and future data are inaccessible. Following \cite{LiuRMM:Learning}, we construct pseudo-CIL tasks entirely from $D_0$, which contains half the full dataset and thus provides sufficient data for meaningful smaller-scale incremental episodes. Each pseudo-task uses a 90/10 train--validation split per phase, enabling reliable reward computation throughout. Each episode proceeds for $N$ phases: at each step, the agent selects $(\alpha_i, \beta_i)$, the CIL model is trained under Eq. \ref{eq:overall-loss}, and the episodic return is the sum of per-phase validation accuracies.

\subsubsection{Learning Algorithm}
\label{sec:learning_alg}
\vspace{-4pt}

\begin{table*}[ht]
\centering
\caption{Performance comparison on average accuracy $ACC$, average forgetting $FRG$ and forward transfer $FWT$ on CIFAR-100.}
\label{table:cifar_results}
\setlength{\tabcolsep}{6pt}
\small
\begin{tabular}{@{}lrrrrrrrrr@{}}
\toprule
 & \multicolumn{3}{c}{CIFAR-100 (N=2)} & \multicolumn{3}{c}{CIFAR-100 (N=5)} & \multicolumn{3}{c}{CIFAR-100 (N=10)} \\
\cmidrule(lr){2-4} \cmidrule(lr){5-7} \cmidrule(lr){8-10}
Method & ACC$\uparrow$ & FRG$\downarrow$ & FWT$\uparrow$ & ACC$\uparrow$ & FRG$\downarrow$ & FWT$\uparrow$ & ACC$\uparrow$ & FRG$\downarrow$ & FWT$\uparrow$ \\
\midrule
JT (Upper Bound)                      & 72.87 & -     & -     & 72.87 & -     & -     & 72.87 & -     & -     \\
\midrule
iCaRL \cite{RebuffiICaRL:Learning}    & 60.32     & 28.54     & -     & 51.96 & 55.29 & -     & 45.40     & 68.26     & -     \\
LUCIR \cite{HouLearningRebalancing}   & 66.94 & 24.55 & 40.18 & 63.93 & 31.92 & 49.42 & 61.24 & 32.13 & 47.68 \\
WA \cite{Zhao2020MaintainingLearning} & 67.2 & - & - & 64.32 & 29.85 & - & 57.77 & \textbf{28.88} & - \\
PodNet \cite{Douillard2020PODNet:Learning} & 67.51 & 30.02 & - & 61.8 & 38.99 & - & 59.22 & 31.35 & - \\
COIL \cite{Zhou2021Co-transportLearning}& 63.50 & 35.34 & - & 55.60 & 38.80 & - & - & - & - \\
OHOCIL \cite{Liu2023OnlineLearning}   & 66.52 & 26.23 & 40.42 & 62.62 & 32.24 & 37.80 & 59.94 & 33.71 & 34.24 \\
\midrule
\textbf{\coolname{} (Ours)}             & \textbf{67.60} & \textbf{23.65} & \textbf{43.78} & \textbf{64.43} & \textbf{28.02} & \textbf{51.77} & \textbf{61.53} & 30.17 & \textbf{49.36} \\
\bottomrule
\end{tabular}
\vspace{-2mm}
\end{table*}

\begin{table}[t]
\centering
\caption{Performance comparison on average accuracy $ACC$, average forgetting $FRG$ and forward transfer $FWT$ on Tiny-ImageNet.}
\label{table:tiny_results}
\setlength{\tabcolsep}{3pt}
\scriptsize
\resizebox{\columnwidth}{!}{%
\begin{tabular}{@{}lrrrrrr@{}}
\toprule
 & \multicolumn{3}{c}{Tiny-ImageNet (N=5)} & \multicolumn{3}{c}{Tiny-ImageNet (N=10)} \\
\cmidrule(lr){2-4} \cmidrule(lr){5-7}
Method & ACC$\uparrow$ & FRG$\downarrow$ & FWT$\uparrow$ & ACC$\uparrow$ & FRG$\downarrow$ & FWT$\uparrow$ \\
\midrule
JT (Upper Bound)                      & 68.78     & -     & -     & 68.78     & -     & -     \\
\midrule
iCaRL \cite{RebuffiICaRL:Learning}    & 37.51    & 31.89     & -     & 42.45     & 32.77     & -     \\
EEIL \cite{castro2018end} & 47.4 & 25.56 & - & 45.2 & 25.91 & - \\
LUCIR \cite{HouLearningRebalancing}   & 56.76 & 15.04 & 56.48 & 53.41 & \textbf{27.60} & 68.12 \\
WA \cite{Zhao2020MaintainingLearning}& 49.28 & 40.85 & - & 38.29 & 37.24 & - \\
PodNet \cite{Douillard2020PODNet:Learning} & 49.9 & 40.99 & - & 44.51 & 34.56 & - \\
FOSTER \cite{Luo2023Class-IncrementalLearning}& 51.85 & 32.35 & - & 44.91 & 36.55 & - \\
OHOCIL \cite{Liu2023OnlineLearning}   & 55.52 & 15.77 & 55.54 & \textbf{54.93}     & 30.65     & 66.5     \\
SCREAM \cite{FENG2024103629} & 52.45 & 29.31 & - & 46.43 & 32.26 & - \\
\midrule
\textbf{\coolname{} (Ours)}             & \textbf{57.68} & \textbf{14.2} & \textbf{61.4} & 47.37 & 39.61 & \textbf{71.75} \\
\bottomrule
\end{tabular}%
}
\vspace{-2mm}
\end{table}

We train the agent with Ape-X DQN \cite{DeepmindDISTRIBUTEDREPLAY}. Since each environment step requires a full CIL training phase, data collection is slow; Ape-X addresses this by distributing collection across parallel actors, each running an independent CIL instance at a different exploration rate. The resulting replay buffer captures a diverse range of experiences, which a central learner consumes in a prioritised fashion.

We use Double DQN~\cite{VanHasseltDoubleQ-learning, vanHasselt2015DeepQ-learning} with $n$-step bootstrap targets \cite{Mnih2016AsynchronousLearning}. The $n$-step return and training loss are:
\begin{equation}
    \label{eq:Gt}
    \begin{aligned}
        G_i &=
        R_{i+1} + \gamma R_{i+2} + \cdots + \gamma^{n-1} R_{i+n} \\
            &\quad + \gamma^n q\!\left(
                S_{i+n},
                \underset{a}{\arg\max}\; q(S_{i+n}, a; \theta),
                \theta^-
            \right),
    \end{aligned}
\end{equation}
\begin{equation}
    \label{eq:agent_train}
    l_i(\theta) = \frac{1}{2}\bigl(G_i - q(S_i, \mathcal{A}_i; \theta)\bigr)^2,
\end{equation}
where $\theta^-$ are the parameters of the slowly updated target network. We set $n$ equal to the number of incremental phases per pseudo-task and $\gamma = 1$, consistent with the finite-horizon MDP in Section \ref{sec:rl_reg}.

\section{Experimental Setup}
\subsection{Datasets}
\vspace{-4pt}

We evaluate on CIFAR-100 (100 classes, $32\times32$ images) and Tiny-ImageNet (200 classes, $64\times64$ images). Following the standard train-from-half protocol, models are first trained on half the classes (50 and 100, respectively), with the remainder introduced over $N\in\{2,5,10\}$ incremental phases.

\subsection{Implementation Details}
\vspace{-4pt}
We use ResNet-32 for CIFAR-100 and ResNet-18 for Tiny-ImageNet, trained with SGD (momentum $0.9$, weight decay $5\times10^{-4}$) and batch size of $128$ for $160$ epochs under a step-decay learning rate schedule, with a fixed memory budget of $2000$ exemplars selected by herding \cite{RebuffiICaRL:Learning}. The RL agent is a three-layer MLP $q(s,a;\theta)$ mapping the four-dimensional state $(A_{i-1}, F_{i-1}, CR_i, IS_i)$ to Q-values over a discrete action set with $\Lambda_\alpha = \Lambda_\beta = \{0.1, 0.5, 1, 1.5, 2, 2.5, 5\}$, optimised with Adam (learning rate $0.01$). We train the policy offline using Ape-X DQN \cite{DeepmindDISTRIBUTEDREPLAY} with 7 parallel actors ($\epsilon$-greedy exploration, $100$ episodes each) and a centralised learner with prioritised replay. To keep offline cost tractable, CIL models during policy training use $20$ epochs per phase on pseudo tasks rather than the $160$ used at deployment. At inference, a single MLP forward pass per phase adds negligible overhead.

\subsection{Evaluation Metrics}
\vspace{-4pt}
We report three metrics: mean accuracy ($ACC$) over all classes seen so far, mean forgetting ($FRG$), and forward transfer ($FWT$), defined as $FWT = A_{i-1,\,\theta_{\text{CL}}} - A_{i,\,\theta_{\text{random}}}$,
where $A_{i,\theta_{\text{random}}}$ is the performance of a randomly initialised model on phase $i$ and $A_{i-1,\theta_{\text{CL}}}$ is that of the CIL model immediately before learning phase $i$. A positive $FWT$ indicates that prior learning facilitates adaptation to new classes.

\section{Results}
\label{sec:results}

\label{sec:acc_vs_frg}

\begin{figure}[t]
    \centering
    \includegraphics[width=0.9\linewidth]{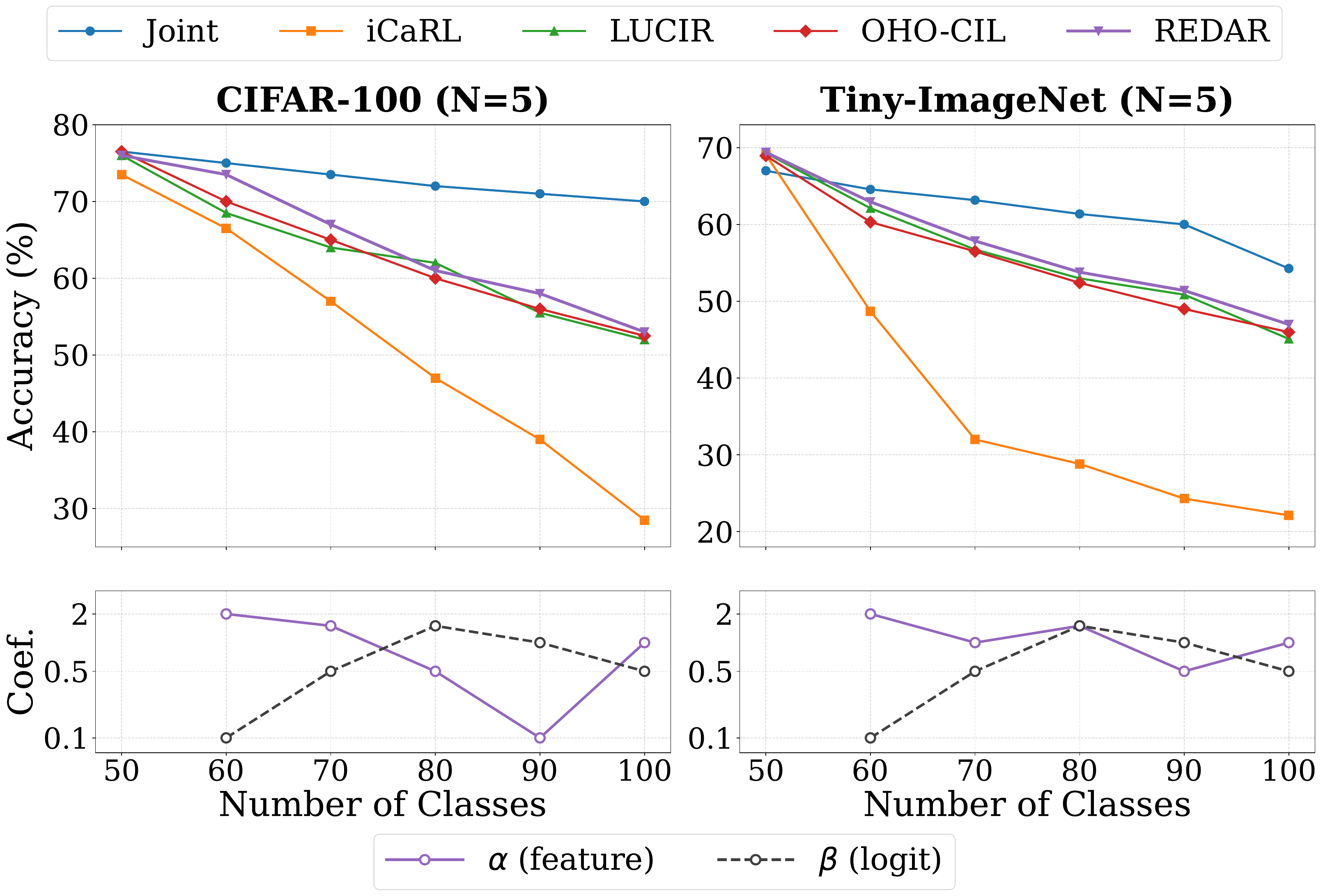}
    \caption{Top: Phase-wise accuracy on CIFAR-100 and Tiny-ImageNet under the $N=5$ incremental setting. Accuracy is reported as the mean over all classes seen after each phase. Bottom: coefficients ($\alpha$, $\beta$) used to train each phase.}
    \label{fig:phasewise}
    \vspace{-2mm}
\end{figure}

Tables~\ref{table:cifar_results} and~\ref{table:tiny_results} report accuracy,
forgetting, and forward transfer across CIFAR-100 (three incremental settings, $N \in
\{2,5,10\}$) and Tiny-ImageNet (two incremental settings, $N \in \{5,10\}$). Figure~\ref{fig:phasewise} shows phase-wise accuracy for $N{=}5$. \coolname{} achieves the best or near-best accuracy on both datasets across most settings, and
consistently leads forward transfer by the widest margin of any method, indicating that
the learned regularisation policy is particularly effective at shaping representations
that generalise to future tasks.


\noindent\textbf{CIFAR-100.} \coolname{} achieves the best or near-best accuracy across
all three incremental settings (67.60\%, 64.43\%, and 61.53\% at $N{=}2,5,10,$
respectively), and leads forward transfer by a substantial margin in every case
(43.78\%, 51.77\%, and 49.36\%, each of which is the highest among all methods), indicating that the
learned regularisation policy consistently produces representations that generalise
well to future tasks rather than merely resisting forgetting on past ones. WA
occasionally reports lower forgetting, most notably at $N{=}10$ (28.88\% vs.\ \coolname{}
30.17\%), but this comes paired with a 3.76-point accuracy deficit relative to \coolname{} 
(57.77\% vs.\ 61.53\%), consistent with a stability-biased regime that suppresses
forgetting by under-acquiring new classes rather than genuinely retaining prior
knowledge more effectively. Taken
together, these results suggest that our adaptive policy resolves the stability-plasticity
trade-off more favourably than static or heuristic baselines, rather than simply trading
one failure mode for another.

\noindent\textbf{Tiny-ImageNet.} At $N{=}5$, \coolname{} leads all three metrics:
highest accuracy (57.68\%), lowest forgetting (14.2\%), and best forward transfer
(61.4\%). At $N{=}10$, OHOCIL and LUCIR attain higher accuracy (54.93\% and 53.41\% vs.\
our 47.37\%) and lower forgetting (30.65\% and 27.60\% vs.\ our 39.61\%), but our forward
transfer remains the strongest overall (71.75\% for \coolname{} vs.\ 68.12\% for LUCIR),
indicating that even where raw retention lags at longer horizons, the learned policy
still yields representations that transfer better to future tasks.

Figure ~\ref{fig:phasewise} (bottom) shows the coefficients selected on each dataset: the policy starts feature-heavy ($\alpha{=}2$, $\beta{=}0.1$) and shifts weight towards logit distillation in the middle phases. Across both datasets, the consistent advantage in forward transfer indicates
\coolname{} shapes representations that generalise better to future tasks than
static or heuristic schedules.


\subsection{Cross-Setting Transfer}
\label{sec:cross_task}
\vspace{-4pt}

\begin{figure}[t]
    \centering
    \includegraphics[width=0.9\linewidth]{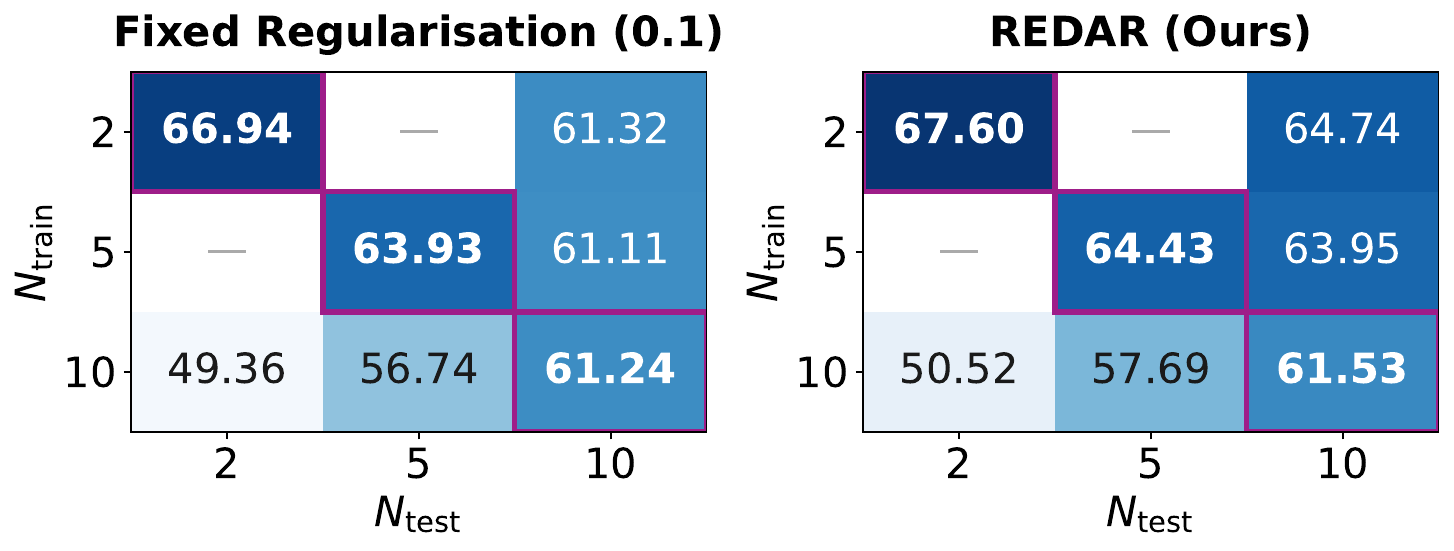}
    \caption{Cross-setting transfer evaluation on CIFAR-100. 
    Each cell reports ACC (\%) for a policy trained on $N_{\text{train}}$ 
    phases and deployed on $N_{\text{test}}$ phases. Diagonal entries 
    (pink outline) correspond to standard in-distribution evaluation. 
    }
    \label{fig:cross-task}
\end{figure}

A key advantage of \coolname{} over fixed hyperparameter methods is that the learned policy is not tied to a specific number of incremental phases. Figure \ref{fig:cross-task} evaluates this by training the policy on $N_{\text{train}}$ phases and deploying it on $N_{\text{test}}$ phases without any retraining. We compare this with a fixed-regularisation hyperparameter setting, representing the best that a static approach can achieve without retuning. 

\coolname{} generalises well across settings without retraining. A policy trained on $N{=}2$ achieves 64.74\% when deployed on $N{=}10$, outperforming fixed hyperparameters tuned specifically for $N{=}10$ (61.24\%). Similarly, a policy trained on $N{=}5$ achieves 63.95\% on $N{=}10$, again surpassing the fixed baseline. In contrast, fixed hyperparameters collapse under cross-setting evaluation: hyperparameters tuned for $N{=}10$ drop to 49.36\% when applied to $N{=}2$, a gap of 17.08\% compared to our policy's 67.60\%. This highlights a fundamental limitation of static regularisation: a single globally tuned coefficient cannot adapt to varying incremental horizons, whereas \coolname{} adjusts dynamically based on the observed state 
at each phase.


\noindent\textbf{Why shorter training horizons transfer to longer ones?} With fewer
phases, each decision carries disproportionate weight on the final return, forcing the
policy to learn a more conservative, generalisable strategy; with more phases, it can
afford early suboptimal decisions and recover later, yielding a more horizon-specific
policy that transfers less well. We attribute this to our state design: conditioning on
$A_{i-1}$, $F_{i-1}$, $CR_i$, and $IS_i$ rather than the phase index lets the policy react
to current training dynamics rather than a fixed schedule, remaining applicable whenever
the observed state matches the training distribution, a phase-agnostic inductive bias
that static hyperparameter selection cannot achieve.

\subsection{Benefit of Adaptive Regularisation}
\label{sec:adapt_reg}
\vspace{-4pt}

\begin{table}[t]
\centering
\small
\caption{Benefit of varying hyperparameters.}
\label{table:hyperparam_abl}
\setlength{\tabcolsep}{6pt}
\begin{tabular}{@{}lccc@{}}
\toprule
Method & ACC$\uparrow$ & FRG$\downarrow$ & FWT$\uparrow$ \\
\midrule
Fixed (0.1) & 63.93 & 31.92 & 49.42 \\
Random Selection                    & 59.84 & 30.28 & 43.16 \\
\textbf{\coolname{} (Ours)}                       & \textbf{64.43} & \textbf{28.02} & \textbf{51.77} \\
\bottomrule
\end{tabular}
\vspace{-2mm}
\end{table}

Table \ref{table:hyperparam_abl} compares \coolname{} against two baselines: a fixed regularisation strength of 0.1 and random selection from the same discrete action space 
$\mathcal{A}$. Random selection serves as a lower bound that isolates the contribution of the learned policy from the benefit of simply varying the regularisation strength across phases. Random selection performs substantially worse than both fixed and learned regularisation (59.84\% ACC, 43.16\% FWT), confirming that varying the regularisation strength alone is not beneficial. The \emph{choice} of strength at each phase matters. \coolname{} outperforms the fixed baseline across all three metrics (64.43\% vs 63.93\% ACC, 28.02\% vs 
31.92\% FRG, 51.77\% vs 49.42\% FWT), demonstrating that the policy learns a meaningful regularisation strategy rather than simply introducing variation. Together, these results confirm that the performance gains of our method stem from the quality of the learned policy, not from the flexibility of the action space alone.

\subsection{Ablation Studies}

\subsubsection{Ablation of State Representation}
\vspace{-4pt}
We validate the contribution of each component in the state representation through a systematic ablation study, with results presented in Table \ref{table:state_abl}. The full state $s_i = (A_{i-1}, F_{i-1}, CR_i, IS_i)$ combines two performance metrics: accuracy $ACC$ and forgetting $FRG$,
with two data properties, class ratio $CR$ and inter-task similarity $IS$. Removing $IS$ while retaining all other components causes a notable drop across all metrics (64.43\% $\to$ 59.97\% ACC, 28.02\% $\to$ 45.46\% FRG), indicating that inter-task similarity is the most influential component. 
This is intuitive: without knowledge of how closely the incoming classes relate to previously seen ones, the agent cannot anticipate whether stronger plasticity or stronger stability is required for the current phase.

Removing $FRG$ while retaining $IS$ has a smaller but still measurable effect, suggesting that forgetting provides complementary signal to accuracy alone. A model with similar overall accuracy may be forgetting old classes at very different rates, and the agent benefits from observing this directly. The full combination of all four components yields the best performance across all metrics, confirming that each component contributes a distinct and non-redundant signal to the policy's decision making.

\begin{table}[t]
    \centering
    \small
    \setlength\tabcolsep{4pt}
    \caption{Ablation results of the state representation}
    \label{table:state_abl}
    \begin{tabular}{ccccccc}
    \hline 
    \noalign{\smallskip}
    \multicolumn{4}{c}{State Representation} & 
    \multicolumn{3}{c}{Metrics} \\
    \noalign{\smallskip}
    \cline{1-4} \cline{5-7} 
    \noalign{\smallskip}
    CR & ACC & FRG & IS & ACC & FRG & FWT \\
    \hline
    \cmark & \cmark & \xmark & \xmark & 62.48 & 37.39 & 45.42 \\
    \cmark & \cmark & \cmark & \xmark & 59.97 & 45.46 & 39.96 \\
    \cmark & \cmark & \xmark & \cmark & 62.49 & 37.44 & 48.82 \\
    \cmark & \cmark & \cmark & \cmark & \textbf{64.43} & \textbf{28.02} & \textbf{51.77} \\
    \hline
    \end{tabular}
    \vspace{-2mm}
\end{table}

\subsubsection{Ablation of Reward Design}
\vspace{-4pt}
\begin{table}[ht]
\begin{center}
\small
\caption{Ablation results of the reward design.}
\label{table:reward_abl}
\begin{tabular}{llll}
\hline\noalign{\smallskip}
Reward Function & ACC & FRG & FWT \\
\hline\noalign{\smallskip}
Learning Effectiveness & 64.27 & 33.38 & 48.66 \\
Validation Accuracy    & \textbf{64.43} & \textbf{28.02} & \textbf{51.77} \\
\hline
\end{tabular}
\end{center}
\vspace{-2mm}
\end{table}
\vspace{-4pt}

We validate the choice of reward function by comparing per-phase validation accuracy with learning effectiveness \cite{SinghLearningMachines}, a composite metric defined as a harmonic-like combination of accuracy $ACC$ and forgetting $FRG$:
\begin{align}
    \label{eq:learn_effec}
    LE = \frac{2}{FRG + \frac{1}{ACC}}.
\end{align}

Results are presented in Table \ref{table:reward_abl}. While learning effectiveness combines both accuracy and forgetting into a single scalar, using it as the reward signal leads to worse performance across all three metrics compared to using validation accuracy directly (64.27\% vs 64.43\% ACC, 33.38\% vs 28.02\% FRG, 48.66\% vs 51.77\% FWT). We attribute this to the composite nature of $LE$: by conflating accuracy and forgetting into a single value, the agent receives an ambiguous training signal that does not clearly distinguish between a model that learns new classes and one that simply retains old knowledge. Per-phase validation accuracy, by contrast, provides a direct and unambiguous measure of the model's current competence across all observed classes, giving the agent cleaner credit assignment at each incremental step.



\section{Conclusions}
\label{sec:conclusion}
\vspace{-4pt}
We present \coolname{}, which shows that viewing regularisation strength in CIL as a controllable
signal, and learning a phase-wise policy for it via reinforcement learning, yields a
more favourable stability--plasticity trade-off than fixed or heuristic schedules. This
adaptivity is most evident as the incremental horizon lengthens: our policy's advantage
over the strongest baseline widens with more phases, and its gains concentrate in
forward transfer, indicating that it does more than resist forgetting on past classes;
it actively shapes representations that generalise to future ones. These findings
suggest that the per-phase stability--plasticity balance is itself a learnable quantity,
rather than something that needs to be fixed in advance or hand-tuned per dataset.
\coolname{} currently operates over a discretised action space for the regularisation
coefficients and is trained offline on a single rehearsal-based backbone; extending it
to continuous control, other CIL backbones (e.g., prompt- or expansion-based methods),
and online adaptation during deployment are natural directions for future work.

\section*{Acknowledgment}
The authors would like to acknowledge support from the CSIRO's Data61 Embodied AI Cluster and partial support from the Australian Research Council (ARC) Discovery Project Grant DP250103634.

{\small
\bibliographystyle{ieee_etal3_unsrt}
\bibliography{egbib}
}

\end{document}